\documentclass[letterpaper, 10 pt, conference]{ieeeconf}  

\IEEEoverridecommandlockouts                              
\usepackage{scalerel}
\usepackage{tikz}
\usetikzlibrary{svg.path}

\definecolor{orcidlogocol}{HTML}{A6CE39}
\tikzset{
  orcidlogo/.pic={
    \fill[orcidlogocol] svg{M256,128c0,70.7-57.3,128-128,128C57.3,256,0,198.7,0,128C0,57.3,57.3,0,128,0C198.7,0,256,57.3,256,128z};
    \fill[white] svg{M86.3,186.2H70.9V79.1h15.4v48.4V186.2z}
                 svg{M108.9,79.1h41.6c39.6,0,57,28.3,57,53.6c0,27.5-21.5,53.6-56.8,53.6h-41.8V79.1z M124.3,172.4h24.5c34.9,0,42.9-26.5,42.9-39.7c0-21.5-13.7-39.7-43.7-39.7h-23.7V172.4z}
                 svg{M88.7,56.8c0,5.5-4.5,10.1-10.1,10.1c-5.6,0-10.1-4.6-10.1-10.1c0-5.6,4.5-10.1,10.1-10.1C84.2,46.7,88.7,51.3,88.7,56.8z};
  }
}

\newcommand\orcidicon[1]{\href{https://orcid.org/#1}{\mbox{\scalerel*{
\begin{tikzpicture}[yscale=-1,transform shape]
\pic{orcidlogo};
\end{tikzpicture}
}{|}}}}

\usepackage{hyperref}

\title{\LARGE \bf
Mixed-Granularity Human-Swarm Interaction}
\author{Jayam Patel$^{1}$\orcidicon{0000-0002-0687-4169}, Yicong Xu$^{2}$ and Carlo Pinciroli$^{1}$\orcidicon{0000-0002-2155-0445} 
\thanks{$^{1}$ Robotics Engineering, Worcester Polytechnic Institute, MA, USA. Email: {\sf \{cpinciroli,jupatel\}@wpi.edu}}%
\thanks{$^{2}$ PTC Inc, MA, USA. Email: {\sf \{yxu\}@ptc.com}}%
}

\usepackage{graphicx}
\usepackage{subcaption}
\usepackage{multirow}
\usepackage{units}
\usepackage{hyperref}

\begin{document}

\maketitle
\thispagestyle{empty}
\pagestyle{empty}

\begin{abstract}
We present an augmented reality human-swarm interface that combines two modalities of interaction: environment-oriented and robot-oriented. The environment-oriented modality allows the user to modify the environment (either virtual or physical) to indicate a goal to attain for the robot swarm. The robot-oriented modality makes it possible to select individual robots to reassign them to other tasks to increase performance or remedy failures. Previous research has concluded that environment-oriented interaction might prove more difficult to grasp for untrained users. In this paper, we report a user study which indicates that, at least in collective transport, environment-oriented interaction is more effective than purely robot-oriented interaction, and that the two combined achieve remarkable efficacy.
\end{abstract}
\section{Introduction}

Robot swarms~\cite{Brambilla2013} are envisioned in large-scale applications for which the direct involvement of humans is undesirable due to high risk of injury or to the impossibility to establish a suitable supply chain. Examples include ocean restoration, planetary exploration~\cite{goldsmith1999book}, deep underground mining~\cite{rubio2012mining}, and forest fires. In these scenarios, autonomy is a necessary condition for the swarm to effectively achieve the mission targets.

However, autonomy is only part of the picture. An equally important aspect of the technology that will realize this vision is that humans must be able to interface with the swarm to issue commands and affect the way these commands are executed during the mission~\cite{kolling_human_2016}.

Despite the importance of the human factor, effective interfaces to interact with robots swarms are currently at their early stages. From a UI/UX standpoint, an interface is effective when (i) it offers a coherent mental model of the system and its purpose and (ii) when the available interactions match this mental model~\cite{Norman2013}. A classical example is the design of windows-based point-and-click interfaces.

Problems (i) and (ii) constitute a considerable hurdle in the design of interfaces for human-swarm interaction. Analogously to the problem of designing swarm algorithms, in human-swarm interaction, a fundamental aspect is the way a user \emph{thinks} a swarm system~\cite{kolling_human_2016}. Broadly speaking, the three primary mental models that have been employed are robot-oriented (i.e., the swarm as a collection of individual robots), swarm-oriented (i.e., the swarm as a coherent unit), and environment-oriented (i.e., modify the environment to specify a goal).

In this paper, we argue that neither approach, alone, is adequate to engage with swarms in an effective way when complex missions must be completed. We argue, instead, that the correct abstraction level must be mixed, and include (at least) an environment-oriented aspect and a robot-oriented aspect. Through environment-oriented primitives, the user can specify high-level goals without directly engaging with the swarm. For example, in a collective transport scenario, the user should dictate where the objects should be moved, rather than assigning tasks to the robots directly. However, at the same time, we recognize that the ability to engage with individual robots can be critical to improve performance. In case of robot failures, for example, a human operator with a global view of the system could be more effective than the swarm itself in reassigning healthy robots to new tasks.

The main contribution of this paper is the first human-swarm interface that enables users to both specify high-level goals and to affect the behavior of individual robots during the mission. For this work, we focused on an inherently collaborative task composed of several phases: collective transport. Using a tablet-based augmented reality application, the user can select the objects to transport and drag them to their intended destination. The swarm then autonomously allocates robots to the task and completes it. During execution, the user can also select individual robots to reassign them to new transport tasks or to replace failed units.

The paper is organized as follows. In Sec.~\ref{sec:relatedwork} we discuss related work on human-swarm interfaces. In Sec.~\ref{sec:methodology} we present our system and its design. In Sec.~\ref{sec:analysis} we report our experimental evaluation, which includes unit tests for the behaviors and a user study on the usability of our application. The paper is concluded in Sec.~\ref{sec:conclusion}.

        
        
        
        
        
    

\section{Related Work} \label{sec:relatedwork}
In reviewing relevant literature, we are mostly concerned with two aspects: the \emph{level of granularity} offered by a specific human-swarm interface, and the \emph{type of tasks} the interface most naturally enables. Regarding the level of granularity, we identify three possible alternatives: \emph{robot-oriented}, \emph{swarm-oriented}, and \emph{environment-oriented}.

Robot-oriented interactions occur when a user must engage with individual robots, e.g., to make them into leaders other robots must follow~\cite{setter_haptic_2015}, to hand-pick robots for a specific task~\cite{natraj_gesturing_2014,nagi_human-swarm_2014,alonso-mora_gesture_2015,kapellmann-zafra_human-robot_2016,gromov_wearable_2016}, or to use a robot as tangible interface for gaming and education~\cite{le_goc_zooids:_2016,ozgur_cellulo:_2017}. The main advantage of these interfaces is the simplicity of their abstraction (the user becomes part of the swarm); however, with collective behaviors in which the user must interact with multiple robots, the downside of this approach is the large amount of information a user must provide to the robots (e.g., in the form of number of user commands per task).

At the opposite side of the spectrum, swarm-oriented interactions occur when a user treats the swarm as a unique entity. This modality of interaction has been demonstrated in navigation tasks, e.g., beacon-based~\cite{bashyal_human_2008}, density-based~\cite{diaz-mercado_distributed_2015}, and waypoint-based~\cite{ayanian_controlling_2014}. The main advantage of swarm-based interaction is that a small number of commands, e.g., the target position, is sufficient to control a large swarm. The price to pay, however, is the lack of fine-grained control on the robots. This makes it impossible to deal with suboptimal task assignment, individual failures, and error cascades.

    \begin{figure}[t]
        \centering
        \includegraphics[width=0.4\textwidth]{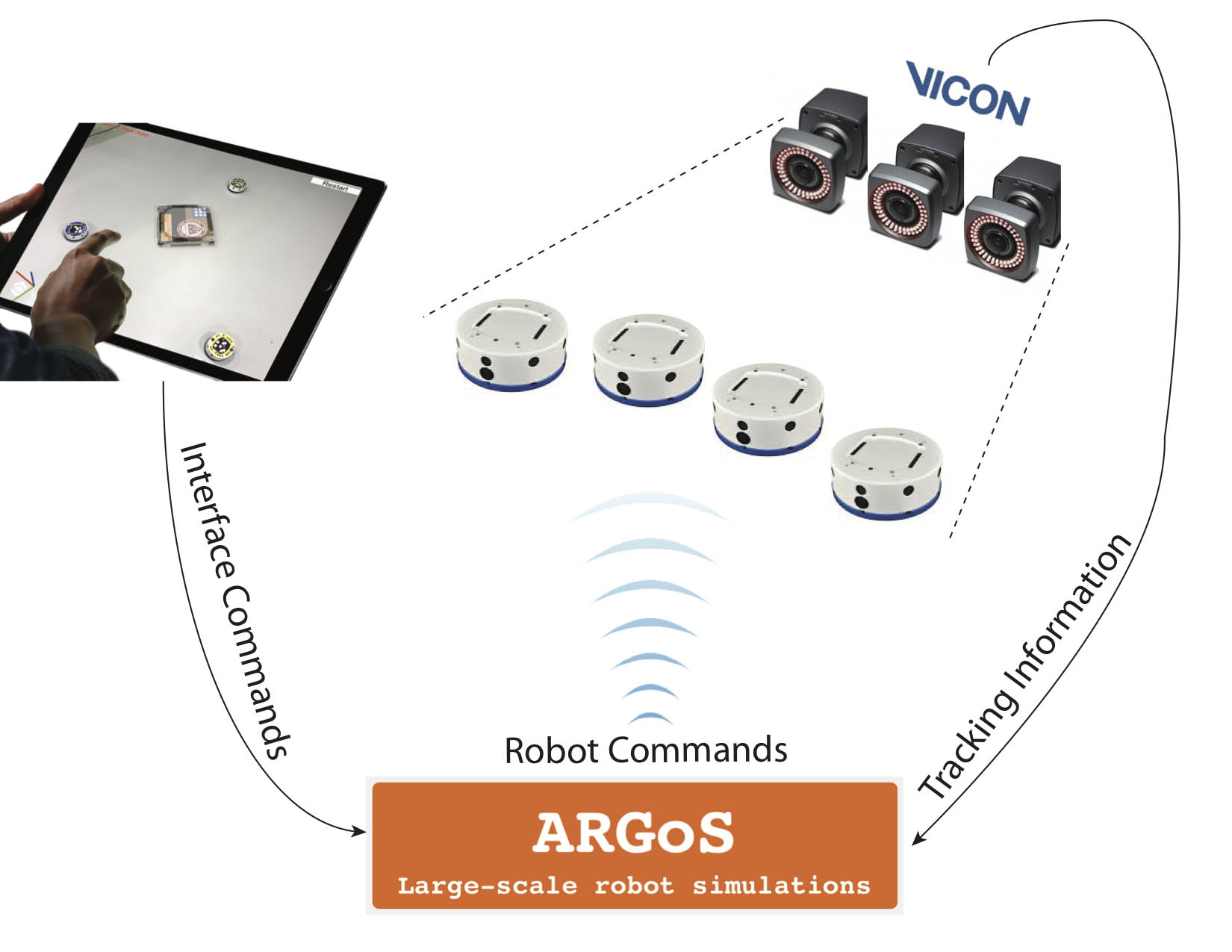}
        \caption{System overview.}
        \label{fig:gbhsi}
    \end{figure}
    
Finally, \emph{environment-oriented} interactions occur when the user does not interact directly with the swarm, but rather performs an environmental modification, either in the real world or in a virtual environment, that the swarm interprets as a new task to perform. Possible examples of this type of interaction include object clustering and sorting, construction, shape formation, and self-assembly of modular structures. The advantage of this modality is that the mental model the user must acquire is very intuitive (describe what you want, rather than how to achieve it) and it is likely to produce concise sets of commands. However, the main disadvantage of environment-oriented interactions is lack of fine-grained control, analogously to what we discussed for swarm-oriented interaction.

Kolling \emph{et al.}~\cite{kolling_human_2013} performed a study that is central to the topic of this paper. They compared two modalities of controlling a swarm, namely robot-oriented and environment-oriented, in a task in which the robots had to diffuse in the environment while avoiding connectivity loss. The robots performed a simple form of foraging, and could be controlled either by direct commands, or by placing attractive beacons in the environment. The conclusions of this study are that environment-oriented interactions are not as effective as robot-oriented interactions, especially when environments are cluttered and many robots are involved.

In this paper, we seek to investigate whether these conclusions depend on the nature of the task (foraging vs.\ collective transport) and whether combining, rather than comparing, environment-oriented and robot-oriented interactions can produce more effective interfaces.

\section{Methodology} \label{sec:methodology}
    
    \subsection{Problem Statement}\label{sec:problemstatement}
    The purpose of this work is to create an intuitive interface to allow a user to interact with a swarm of robots at two levels: the goal and the individual robots. Through the interface, the user should be able to create swarm-level goals, affect the behavior of individual robots, and monitor the progress of the swarm.
    
    To highlight the collaborative aspect of the task that the swarm must accomplish, we opted to focus on a specific scenario---collective transport. Collective transport entails several phases: assignment of robots to the task, approaching the object, navigating, and performing correcting maneuvers when necessary. Hence, we consider it a suitable testbed for a mixed-granularity interface.
    
    

    \subsection{System Overview}
        
        \begin{figure}[t]
            \centering
            \includegraphics[width=0.4\textwidth]{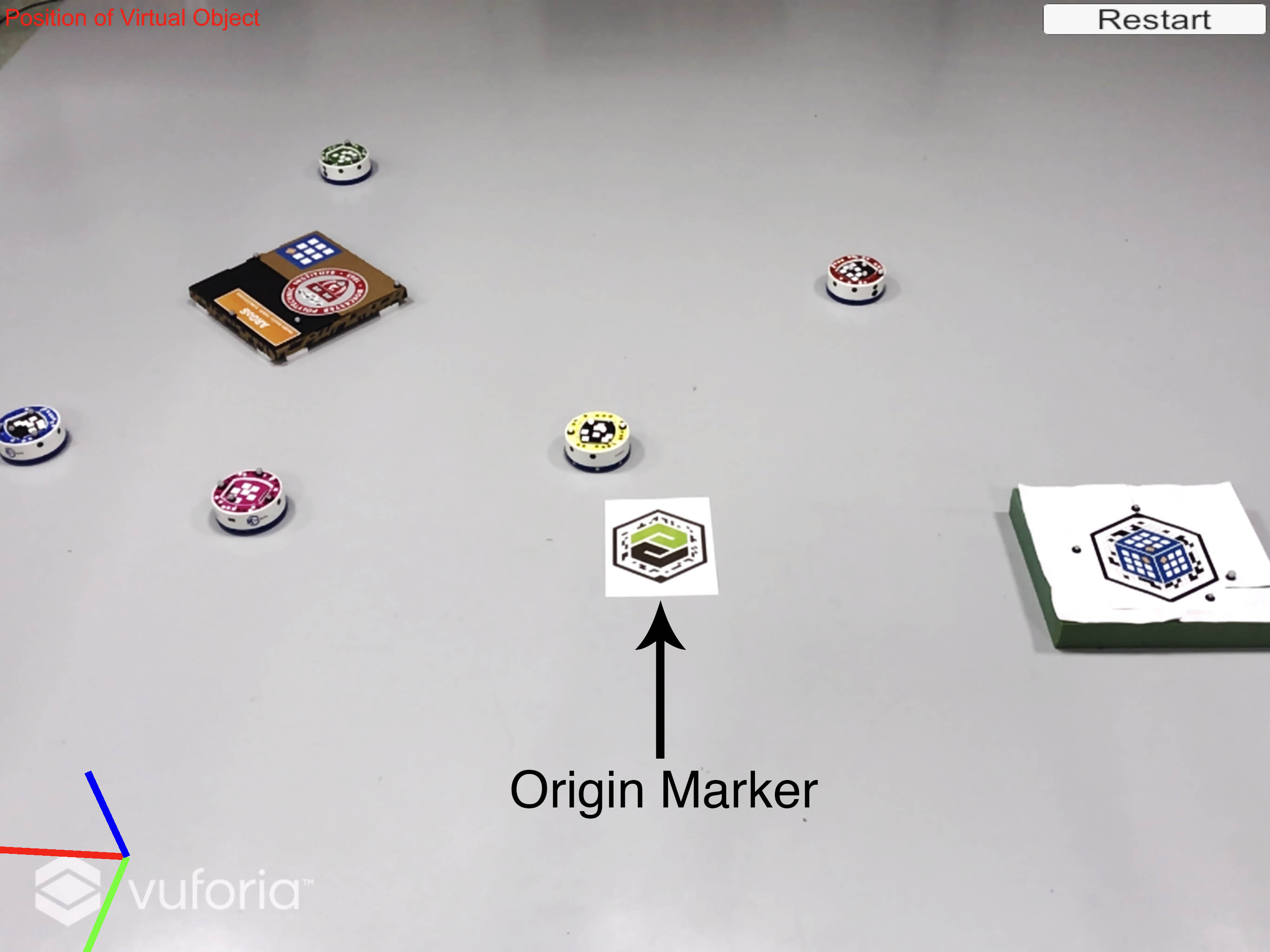}
            \caption{Screenshot of the AR Interface running on an iPad. The overlaid black arrow indicates the origin marker for initializing the coordinate frame of the interface.}%
            \label{fig:app}
        \end{figure}
        
        The system (a diagram of which is reported in Fig.~\ref{fig:gbhsi}) comprises four components:
        \begin{enumerate}
            \item An augmented reality interface implemented as an app for a hand-held device;
            \item A swarm of robots performing collective transport;
            \item A 10-camera VICON motion tracking system, which monitors the position of the robots and of the objects being transported; and
            \item ARGoS~\cite{Pinciroli:SI2012}, a multi-robot simulator that we modified to act as a software glue for the overall system.
        \end{enumerate}
        
        The information flow starts at the hand-held device, when the user defines a new transport goal or designates a new position for a robot. The command is then transmitted to ARGoS, which processes it and generates high-level motion goals for the robots. ARGoS communicates the motion goals to the robots, and the latter execute the goals.
        
        To make this information flow possible through ARGoS, we realized a series of extensions. The most important is a new type of physics engine that, instead of calculating values from a numerical model, uses the positional information generated by the motion tracking system. In this context, ARGoS ceases to be a simulator and it becomes a middleware.
        
    
     \subsection{User Interface}
        
            \begin{figure}[t]
                \centering
                \begin{subfigure}[t]{0.23\textwidth}
                    \includegraphics[width=\textwidth]{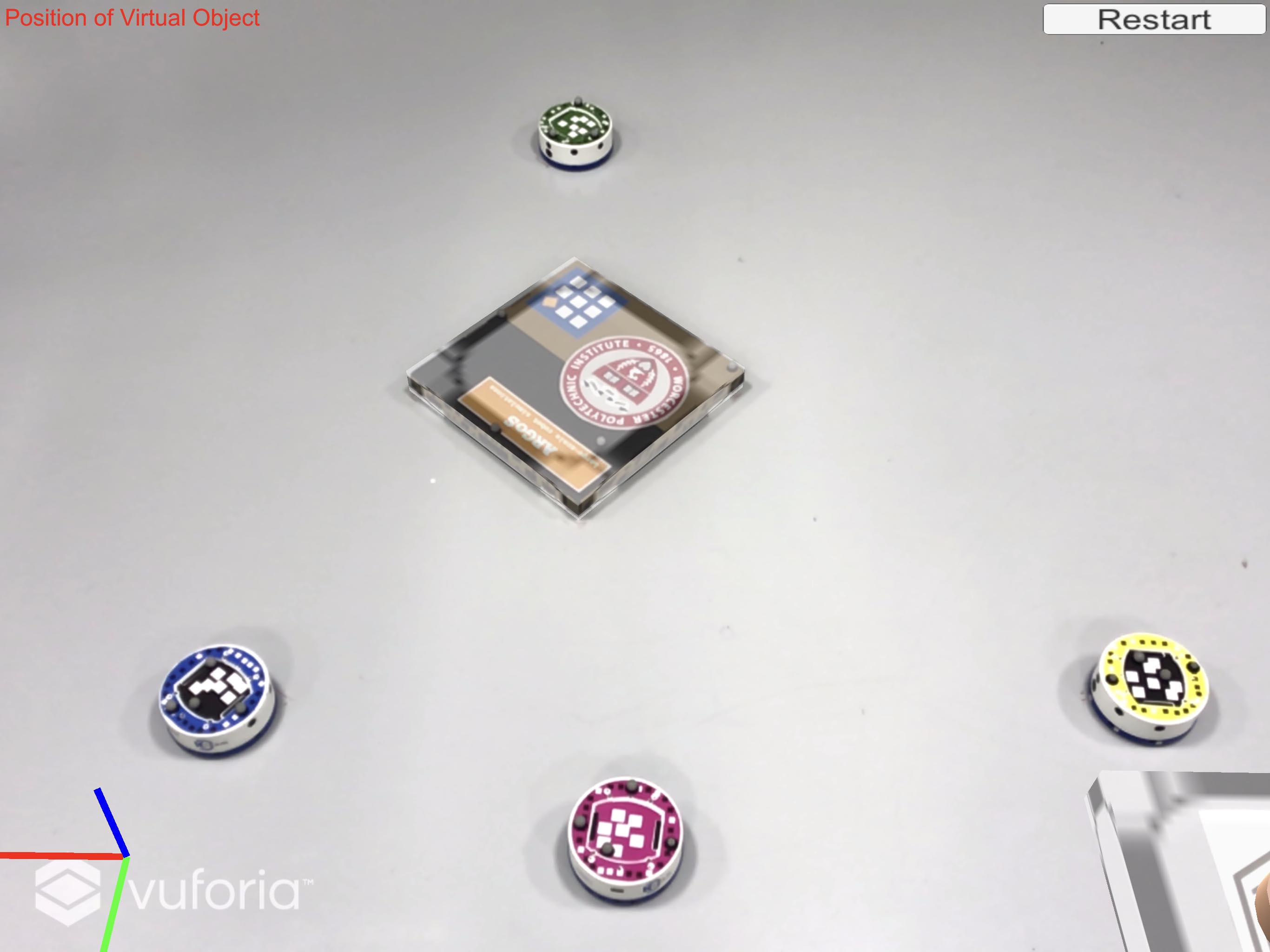}
                    \caption{Object recognition}
                    \label{fig:mode1a}
                \end{subfigure}
                \begin{subfigure}[t]{0.23\textwidth}
                    \includegraphics[width=\textwidth]{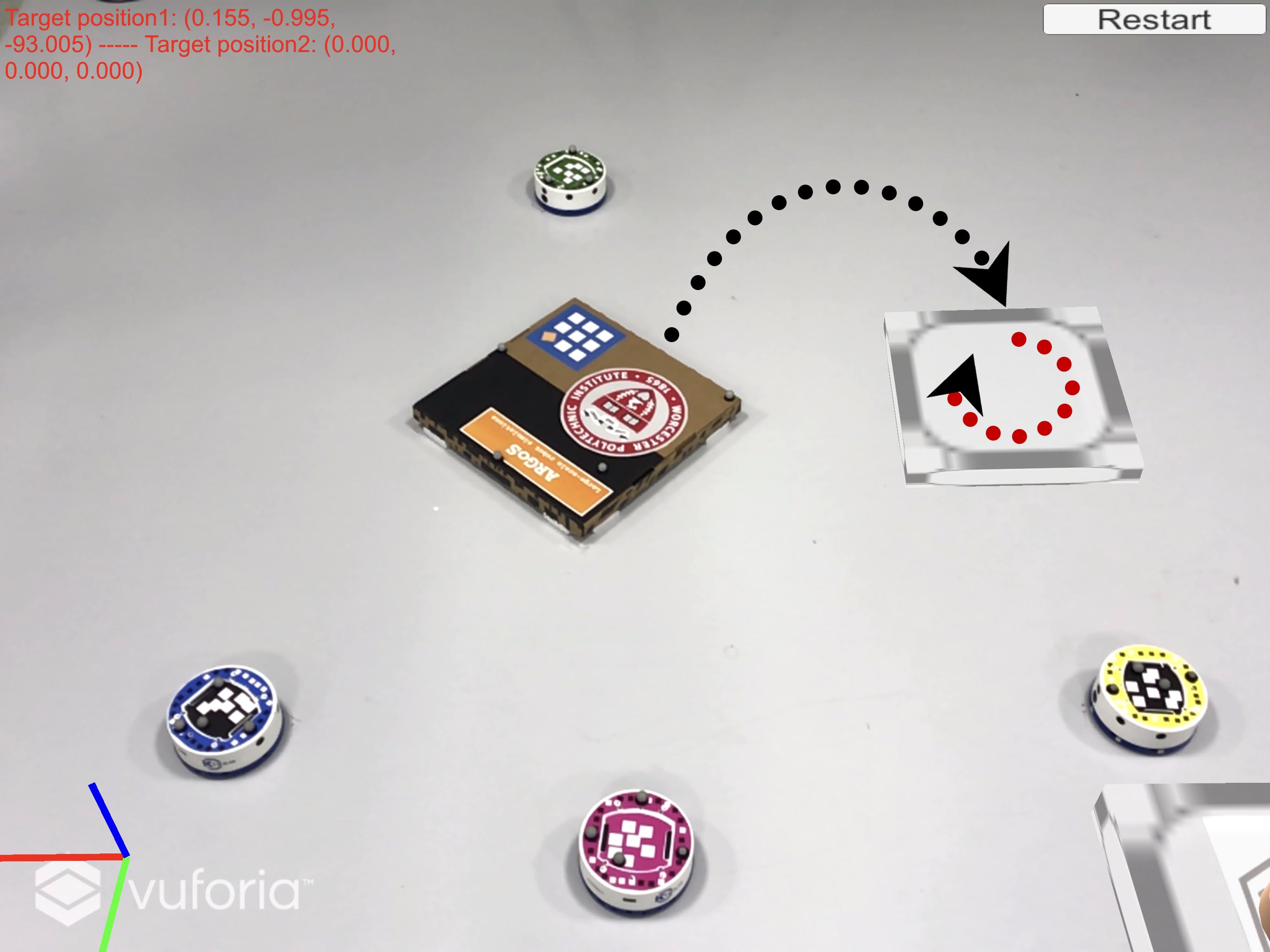}
                    \caption{New Goal Defined}
                    \label{fig:mode1b}
                \end{subfigure}
                \begin{subfigure}[t]{0.23\textwidth}
                    \includegraphics[width=\textwidth]{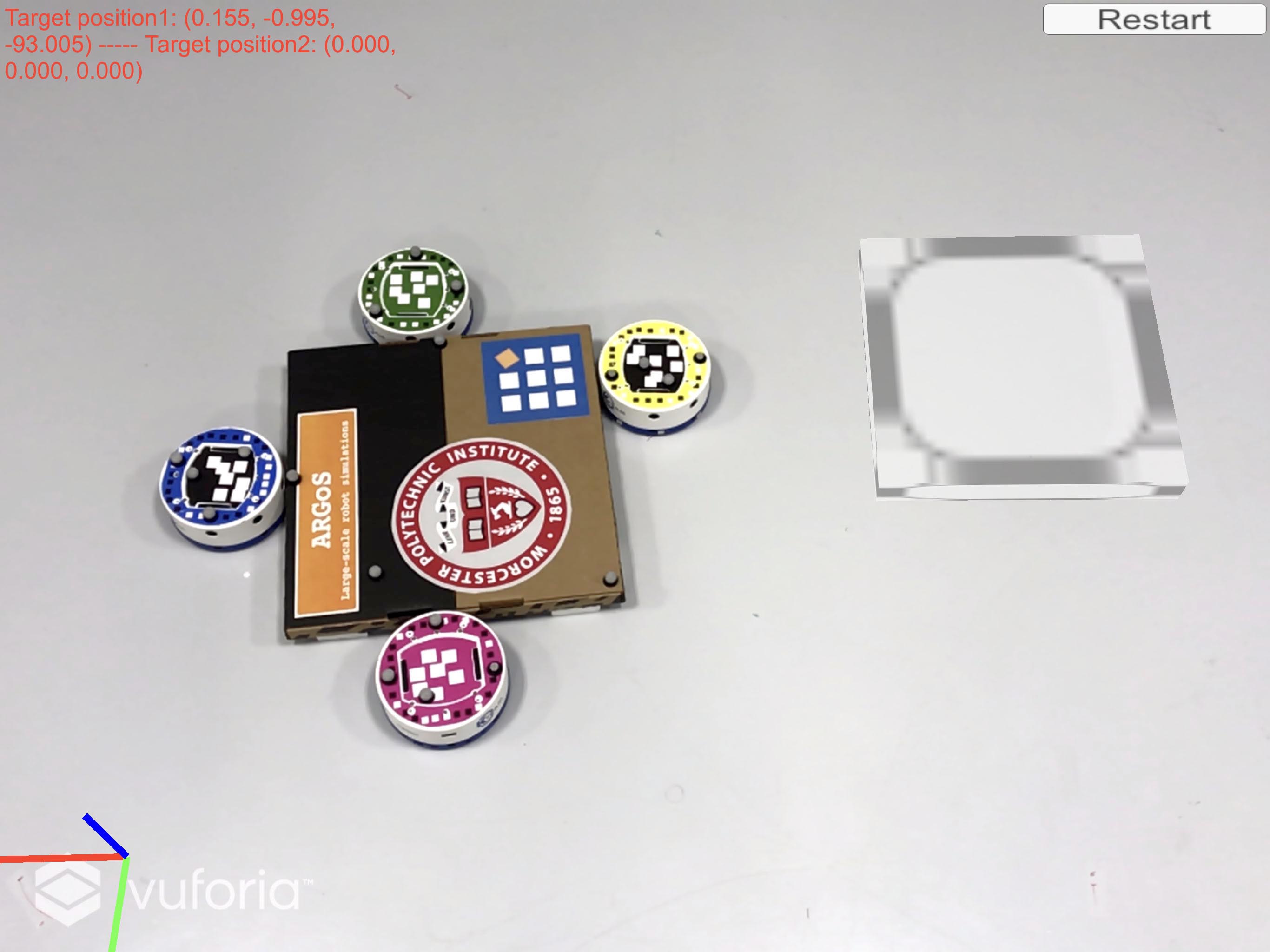}
                    \caption{Robots approach and push}
                    \label{fig:mode1c}
                \end{subfigure}
                \begin{subfigure}[t]{0.23\textwidth}
                    \includegraphics[width=\textwidth]{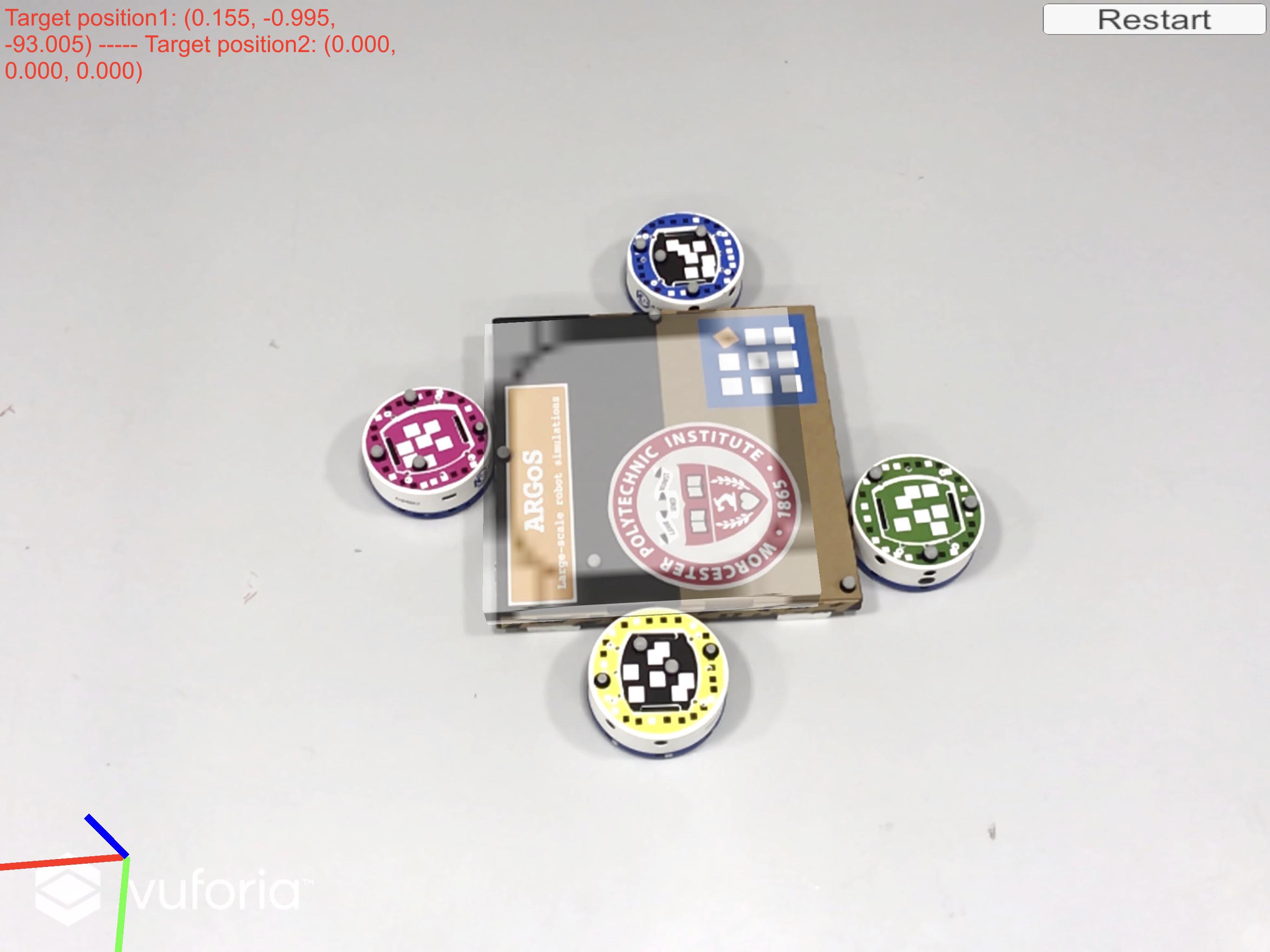}
                    \caption{Transport complete}
                    \label{fig:mode1d}
                \end{subfigure}
                \caption{Goal manipulation by interacting with the virtual object through the interface. The overlaid dotted black arrow indicates the one-finger swipe gesture used to move the virtual object and the overlaid red dotted arrow indicates the two-finger rotation gesture.}\label{fig:mode1}
            \end{figure}
            \begin{figure}[t]
                \centering
                \begin{subfigure}[t]{0.23\textwidth}
                    \includegraphics[width=\textwidth]{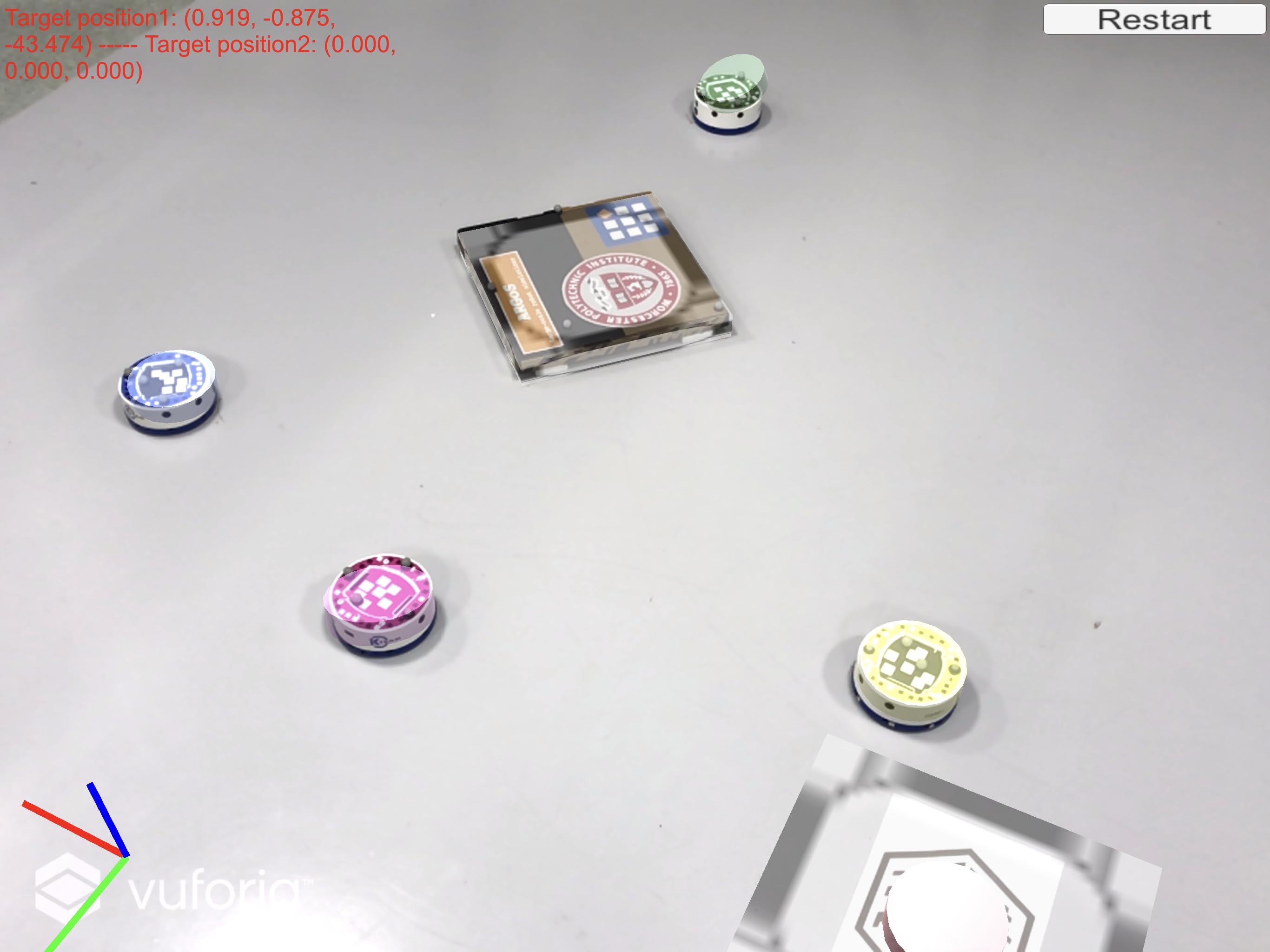}
                    \caption{Robot recognition}
                    \label{fig:mode2a}
                \end{subfigure}
                \begin{subfigure}[t]{0.23\textwidth}
                    \includegraphics[width=\textwidth]{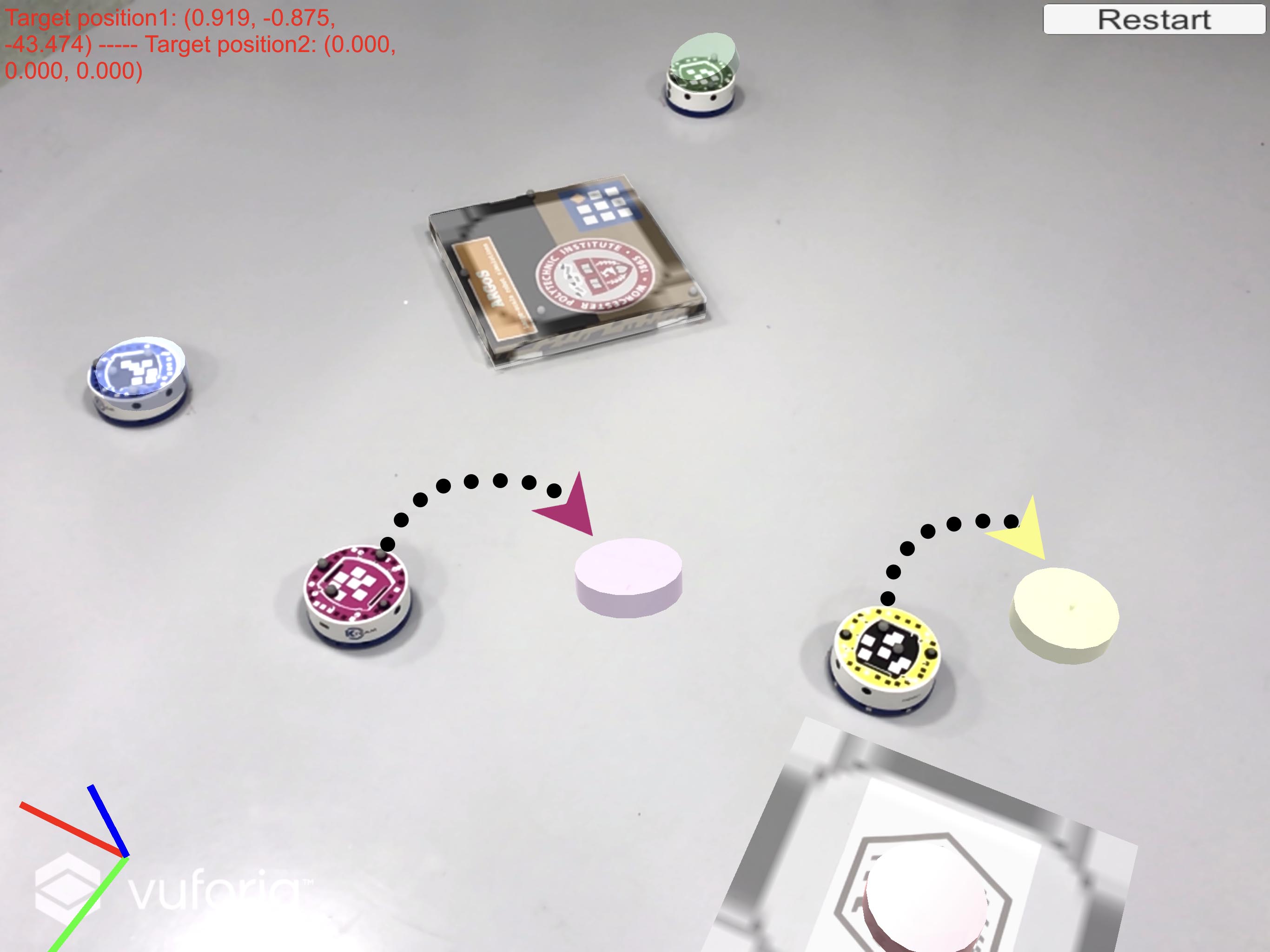}
                    \caption{New robot position}
                    \label{fig:mode2b}
                \end{subfigure}
                \begin{subfigure}[t]{0.23\textwidth}
                    \includegraphics[width=\textwidth]{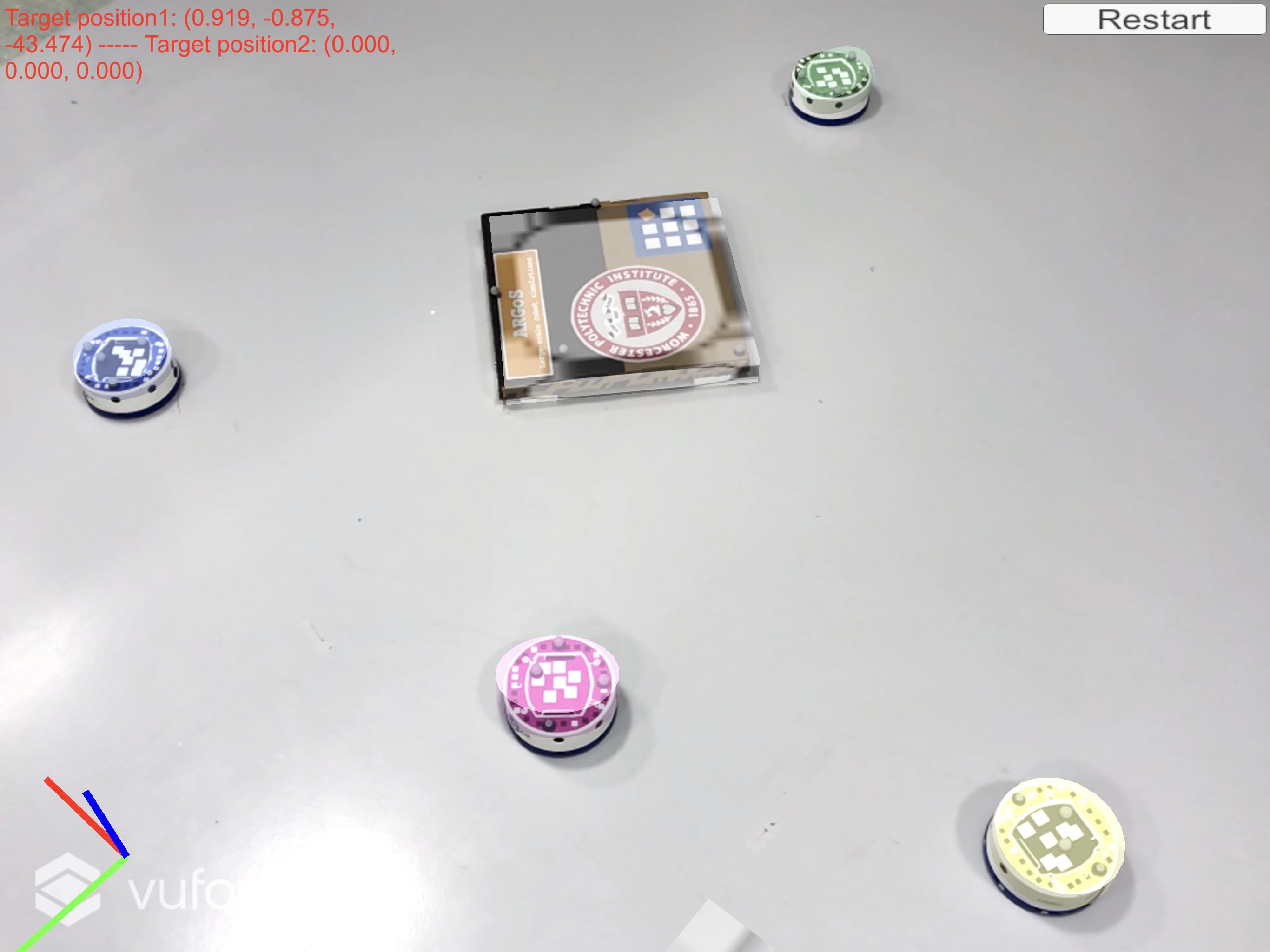}
                    \caption{Robots re-positioned}
                    \label{fig:mode2c}
                \end{subfigure}
                \caption{Robot manipulation by interacting with the virtual robots through the interface. The overlaid dotted black arrow indicates the one-finger swipe gesture to move the virtual object and the arrowhead color indicates the moved virtual robots.}\label{fig:mode2}
            \end{figure}        
        \textbf{Human-Swarm Interaction App.} The interaction between the user and the robot swarm happens through the Augmented Reality (AR) application installed on an iOS 9+ hand-held device. The application can recognize objects and robots. Once recognized, the app overlays a physical entity (robot or object) with a virtual object. The user can specify the desired translation and rotation of the physical object by manipulating its virtual counterpart. A virtual object is translated using a one-finger swipe and rotated using a two-finger twist gesture. The manipulation of a virtual object happens in three touch phases; \emph{start}, \emph{move}, and \emph{end}. At the start of the touch phase, the app selects the virtual object intersecting with the touch point. The move phase records the motion gestures input by the user. In the end phase, the app sends the final pose of the virtual object to ARGoS. Figure \ref{fig:app} shows the screenshot of the AR application. The top-left corner of the application displays the desired goal position; the bottom-left corner displays the current reference frame based on the location of the device with respect to the origin marker. The origin marker can be any image, as long as the app can uniquely identify it.
        
        
        \textbf{Augmented Reality Engine.}
        To realize the app we employed Vuforia~\cite{vuforia}, a well-known software development kit for augmented reality applications. Vuforia uses fiducial markers for recognition and tracking of physical objects in real time. Vuforia provides simultaneous tracking of 5 image targets and 2 object targets. Vuforia can track a \unit[0.2]{m}-wide target from a distance of \unit[2]{m}, but the actual readings may vary based on light conditions, camera resolution, camera focus, and features of the fiducial marker. To develop our app, we integrated Vuforia with the Unity Game Engine, which natively supports several hand-held devices.
        
    \subsection{Interaction Modes}    
        \textbf{Goal Manipulation.} As explained, when the app recognizes an object, it overlays a virtual object over it. This virtual object can be manipulated to generate the desired goal pose for the physical object. Multiple objects can be manipulated through this gesture and the respective robot swarms can transport the objects in parallel. If all robots are busy transporting other objects, the app queues the request and waits for the completion of ongoing tasks. Fig.~\ref{fig:mode1a} shows a detected object overlaid with a virtual object. Fig.~\ref{fig:mode1b} illustrates the manipulation of the virtual object.

        \textbf{Robot Manipulation.}
            When the app recognizes a robot, it overlays a virtual colored marker over it. The color of the virtual marker resembles the color of the physical markers glued to the corresponding robot. Once the user gestures a new pose, other robots belonging to the same swarm as the selected robot freeze until the latter achieves its new pose. During freeze time, multiple robots can be manipulated in parallel. Fig.~\ref{fig:mode2a} shows the identified robots with virtual markers overlaid. Fig.~\ref{fig:mode2b} shows the manipulation of a virtual marker that encodes the new robot pose.
            
\begin{figure}[t]
                \centering
                \begin{subfigure}[t]{0.23\textwidth}
                    \includegraphics[width=\textwidth]{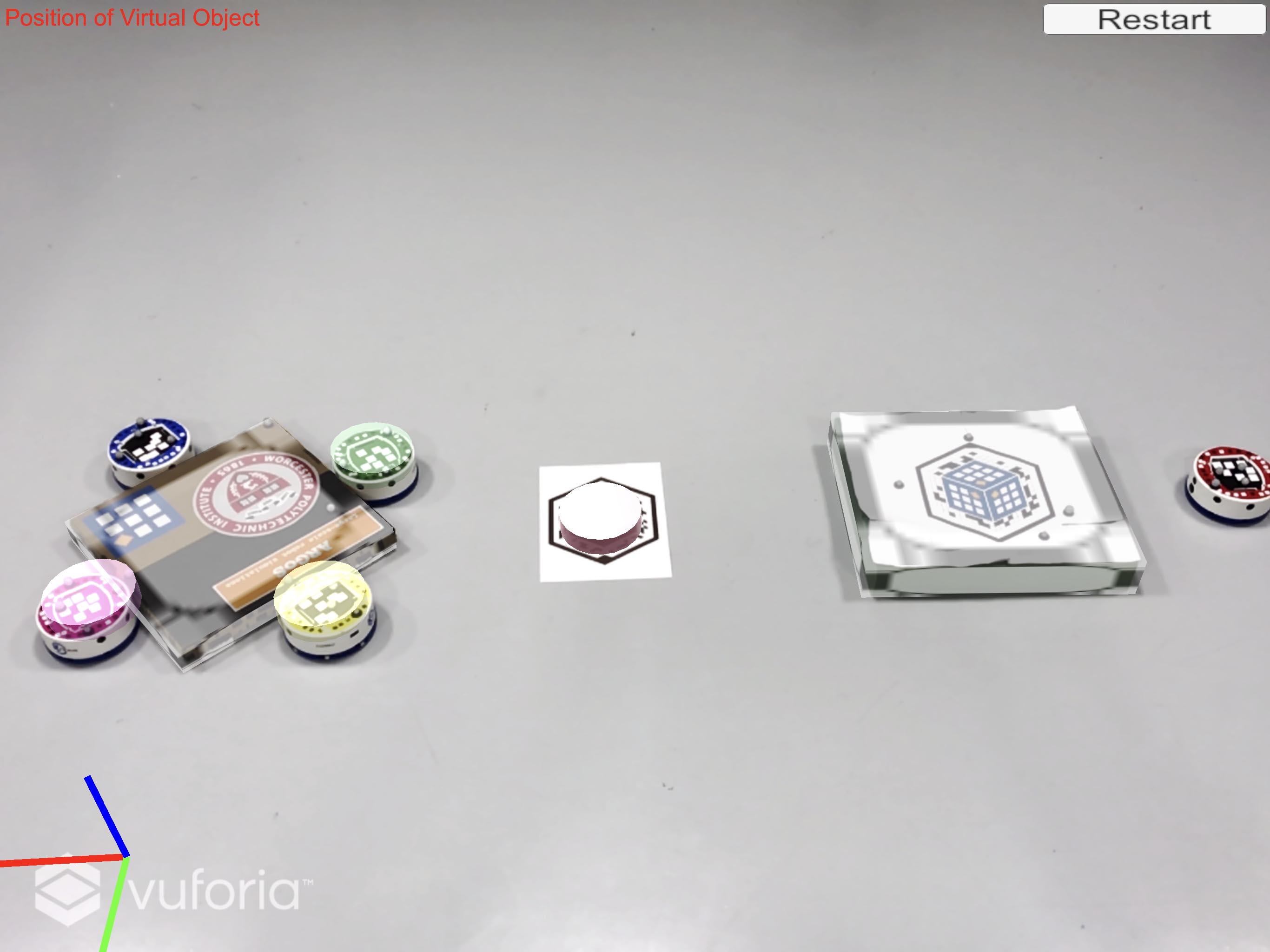}
                    \caption{Two object tasks}
                    \label{fig:mode3a}
                \end{subfigure}
                \begin{subfigure}[t]{0.23\textwidth}
                    \includegraphics[width=\textwidth]{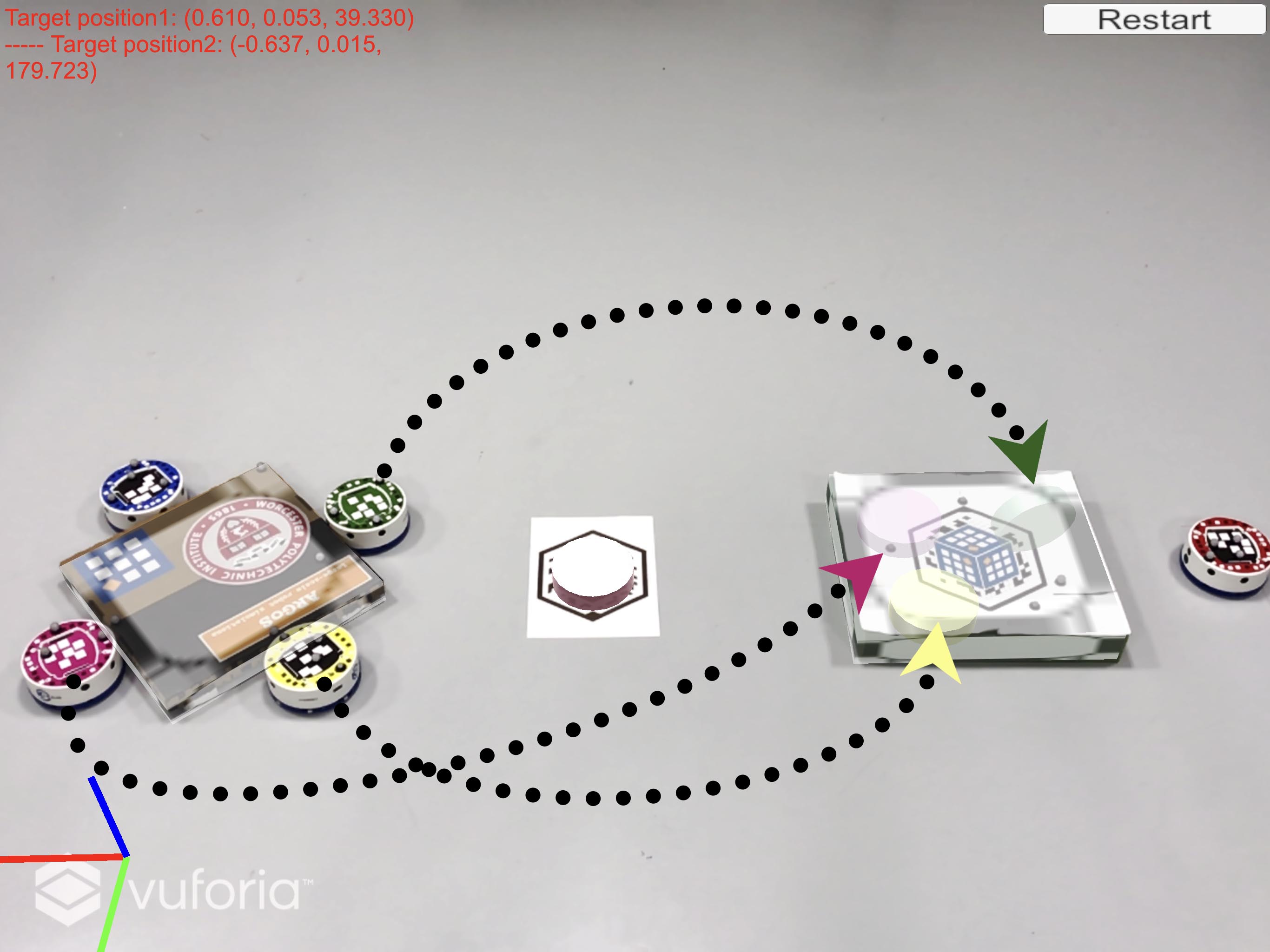}
                    \caption{Team reassignment gesture}
                    \label{fig:mode3b}
                \end{subfigure}
                \begin{subfigure}[t]{0.23\textwidth}
                    \includegraphics[width=\textwidth]{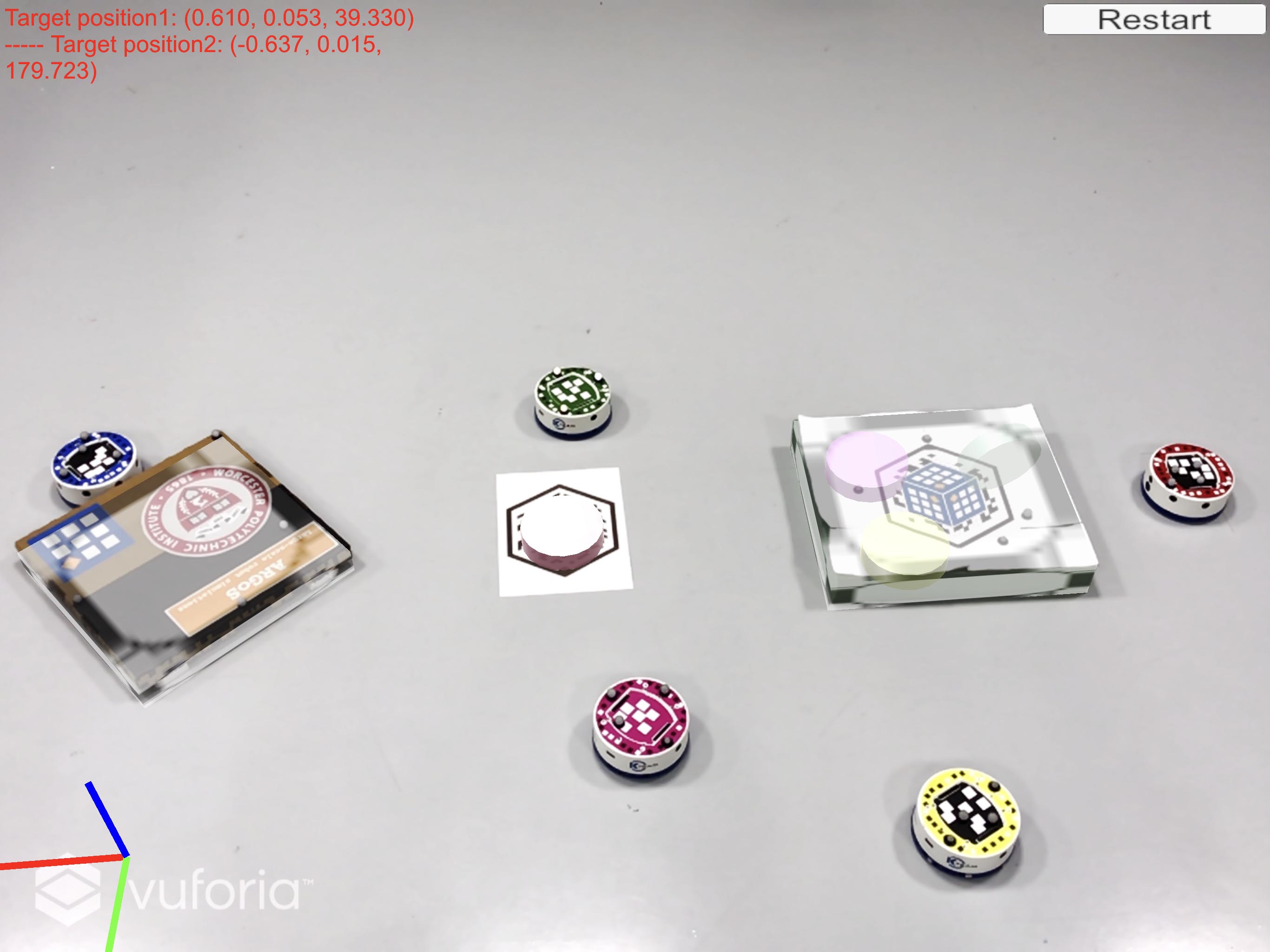}
                    \caption{Robots rearrange}
                    \label{fig:mode3c}
                \end{subfigure}
                \begin{subfigure}[t]{0.23\textwidth}
                    \includegraphics[width=\textwidth]{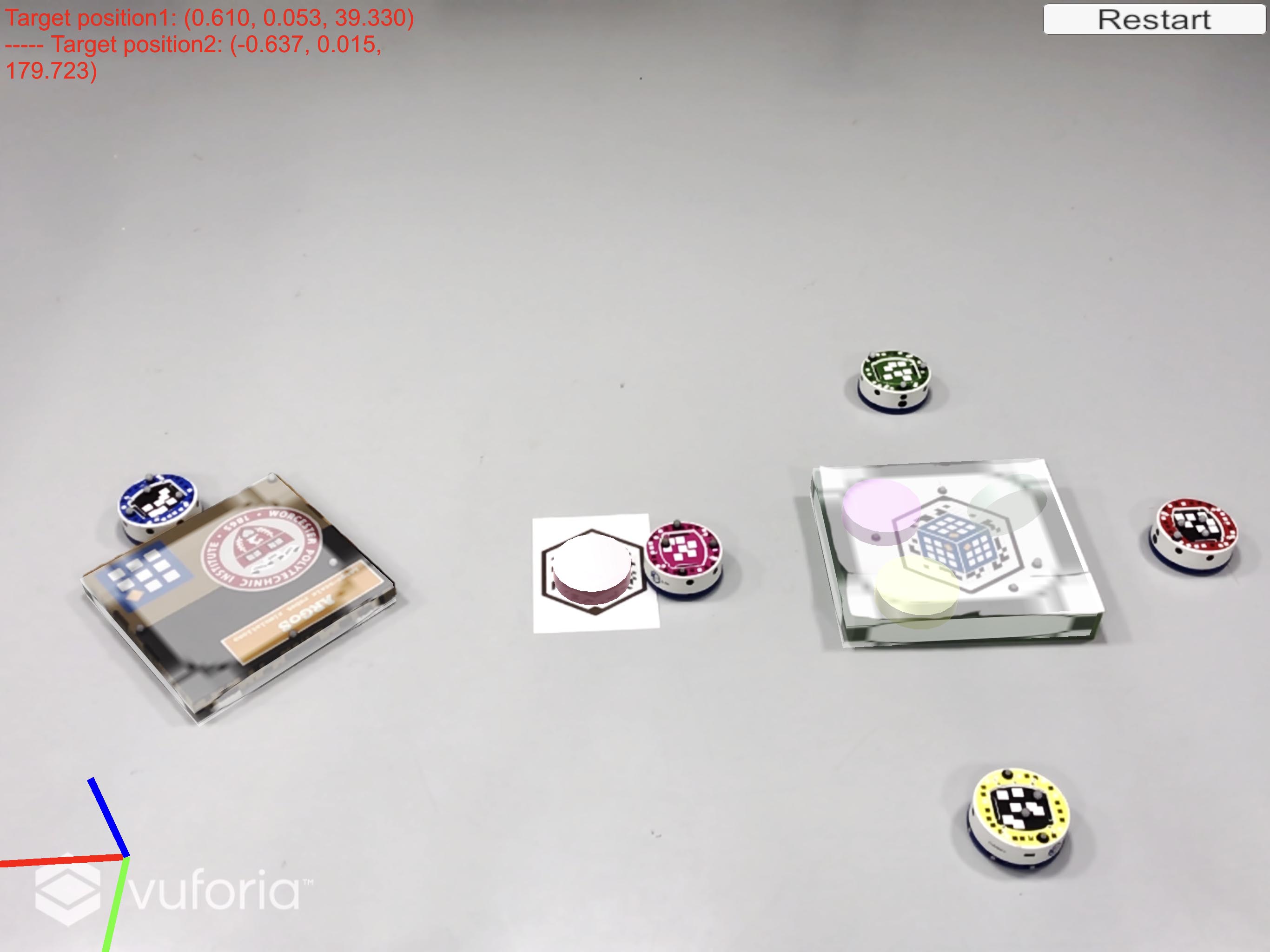}
                    \caption{Team reassigned}
                    \label{fig:mode3d}
                \end{subfigure}
                \caption{Team reassignment through the interface to complete the task of moving object. The overlaid dotted black arrow indicates the one-finger swipe gesture to move the virtual object and the red dotted arrow indicates the two-finger rotation gesture.}\label{fig:mode3}
            \end{figure}

        \textbf{Team Reassignment.}
            The virtual robots can be selected, moved, and overlapped with the virtual objects to reassign them to a new transport task. This interface mode could be useful when one task has insufficient robots, for instance, because the object to be transported requires extra effort. Fig.~\ref{fig:mode3} shows two swarms of robots performing collective transport while the user rearranges the robots and assigns them to an incomplete task.

    \subsection{Collective Transport}
        
        \begin{figure}[t]
            \centering
            \includegraphics[width=0.4\textwidth]{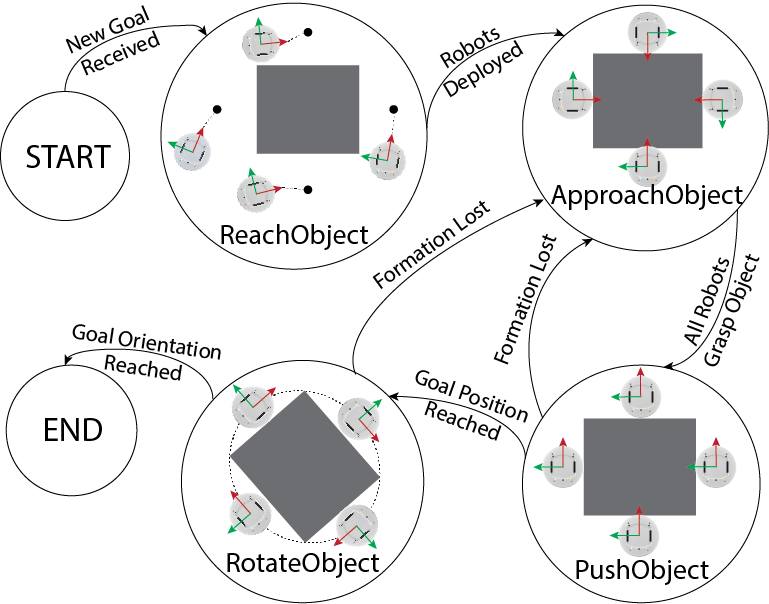}
            \caption{Collective transport state machine}
            \label{fig:statemachine}
        \end{figure}
        
        The collective transport behavior is structured into a state machine, as shown in Fig.~\ref{fig:statemachine}. Since the focus of our work is on the user interface, we kept the transport behavior as simple as possible, but sufficiently effective to act as a meaningful use case for our interface. Designing a more complex, or more decentralized transport behavior is beyond the scope of this work. The states in the FSM are described next.
        
        \textbf{ReachObject.} The robots calculate the direction vector to the assigned object, and then navigate while avoiding obstacles. Depending on the number of robots assigned to an object, the deployment positions are generated in a circular fashion around the object, resulting in a swarm of robots caging their assigned object. New deployment positions are generated every time a robot is removed from or added to a team. The state comes to an end once all the robots reach their deployment positions.

        \textbf{ApproachObject.} From the deployment positions, the robots move towards the centroid of the object. The state ends when all the robots are in contact with the object.
        
        \textbf{PushObject.} The robots rotate in place facing the direction of the goal and start moving at the same speed towards the goal. In particular, the front robot adjusts its speed while maintaining a specific distance from the object. This feature prevents the robots from breaking formation while pushing the object. In case a robot loses the formation, the swarm re-deploys and re-approaches the object. The state ends successfully when the object reaches the goal position. 
        
        \textbf{RotateObject.} All robots rotate in place facing outwards in a circular manner, and move along a circle to rotate the object. In case a robot breaks the formation, the swarm re-deploys and re-approaches the object. The state ends successfully when the object's orientation is within an acceptable value with the respect to the goal orientation.  
        
        
        

            
        

\section{Experimental Analysis} \label{sec:analysis}
    In this section, we analyze the performance of the collective transport behavior and the usability of the app interface assessed through a user study.
    
        \begin{figure}[t]
            \centering
            \includegraphics[width=0.4\textwidth]{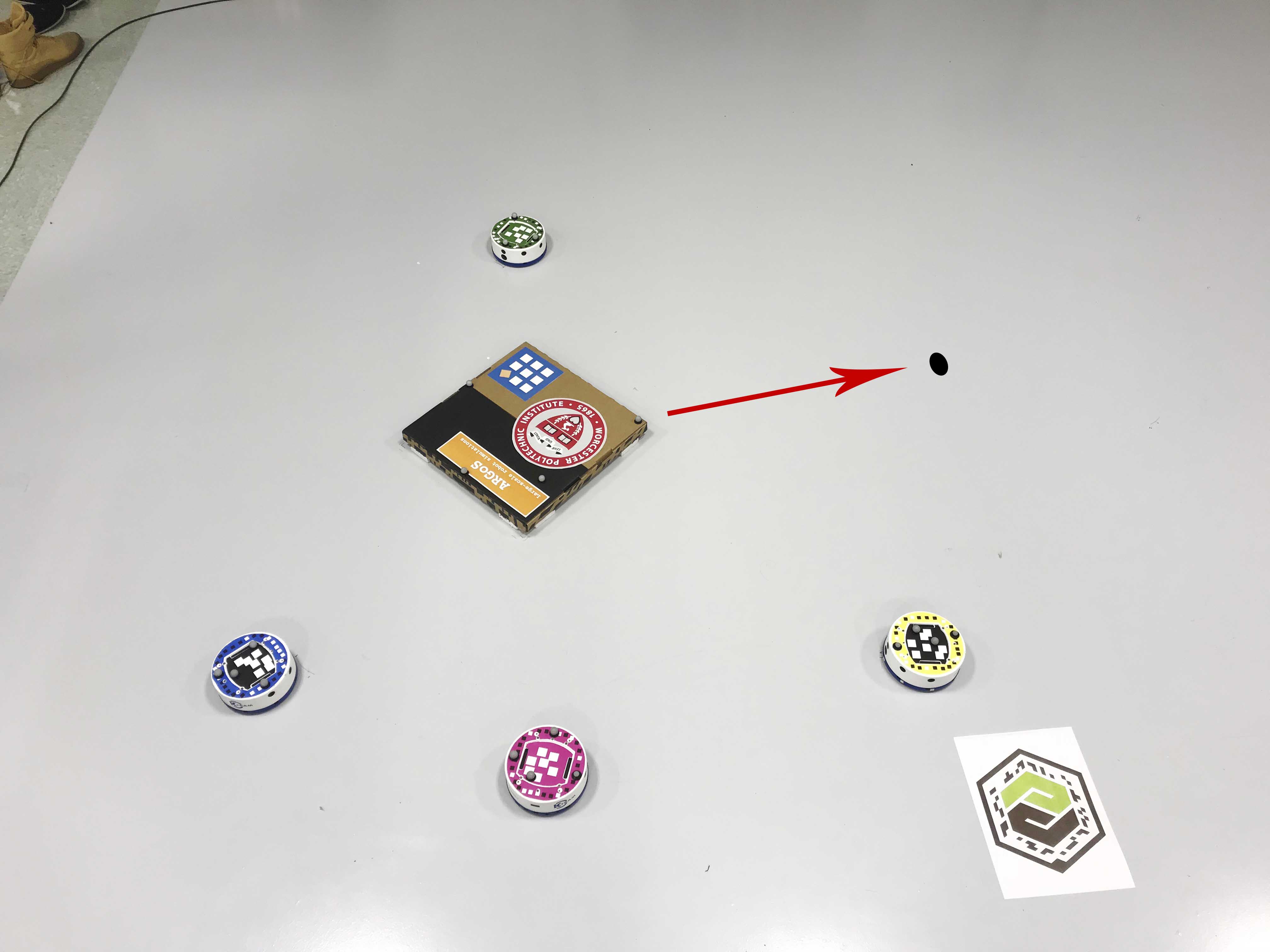}
            \caption{The setup of Experiment 1. The overlaid red arrow and black point indicate the direction and position of the object to be transported by the robots.}
            \label{fig:exp1}
        \end{figure}
        
    \subsection{Transport Behavior Analysis}
        
    As a preparatory step towards the user study, we characterized the performance of the transport behavior to ensure that the task completion rate was within acceptable bounds. We also evaluated the need for human intervention in completing the task when there are insufficient robots in a team.
    
    The aim of Experiment 1 was to transport one object to a predefined target pose using a team of four robots. In Experiment 2, the robot team had to transport two objects to a predefined goal pose using five robots. In Experiment 2 a human was present in case robots had to be reassigned from a task to another.
        
        
        \textbf{Experiment 1: Setup.} We performed 10 consecutive trials of collective transport of an object with four robots (Fig.~\ref{fig:exp1}) to achieve a predefined pose ($x = \unit[0]{m}$, $y = \unit[-1]{m}$, $\theta = 152^{\circ}$). To minimize the effect of spurious failures not related to our algorithm, we selected the target pose according to (i) the coverage of the motion capture system across the arena and (ii) the presence of floor irregularities.
        

	    \textbf{Experiment 1: Results.} Statistics about the recorded final positions and orientations across the 10 trials are reported in Table~\ref{tab:exp1}. The position errors are in meters and the orientation error is in degrees. The data shows that the collective transport behavior is very efficient in moving the object, as errors are on average in the order of cm and faction of a degree. The average completion time, about 2 minutes, is also adequate for tests involving untrained users.
	    
	    \begin{table}[t]
            \centering
            \caption{Statistics on data collected from Experiment 1}
            \begin{tabular}{c||c|c|c|c}
            \hline
            Data type (unit) & Min & Max & Median & Mean \\
            \hline
                Error on $x$ (m) &	            -0.0635 &	-0.0443 &	-0.0571 &	-0.0563 \\
                Error on $y$ (m) &	            -0.0538 &	-0.0325 &	-0.0461 &	-0.0450 \\
                Error on $\theta$ ($^{\circ}$) &	    -0.0290 &	 1.3250 &	1.1085 &	1.0040 \\
                Completion Time (s) &	        121.2   &	 148.9  &	123.1 &	    128.5       \\
            \hline
            \end{tabular}
            \label{tab:exp1}
        \end{table}

	    
	    \begin{figure}[t]
            \centering
            \includegraphics[width=0.4\textwidth]{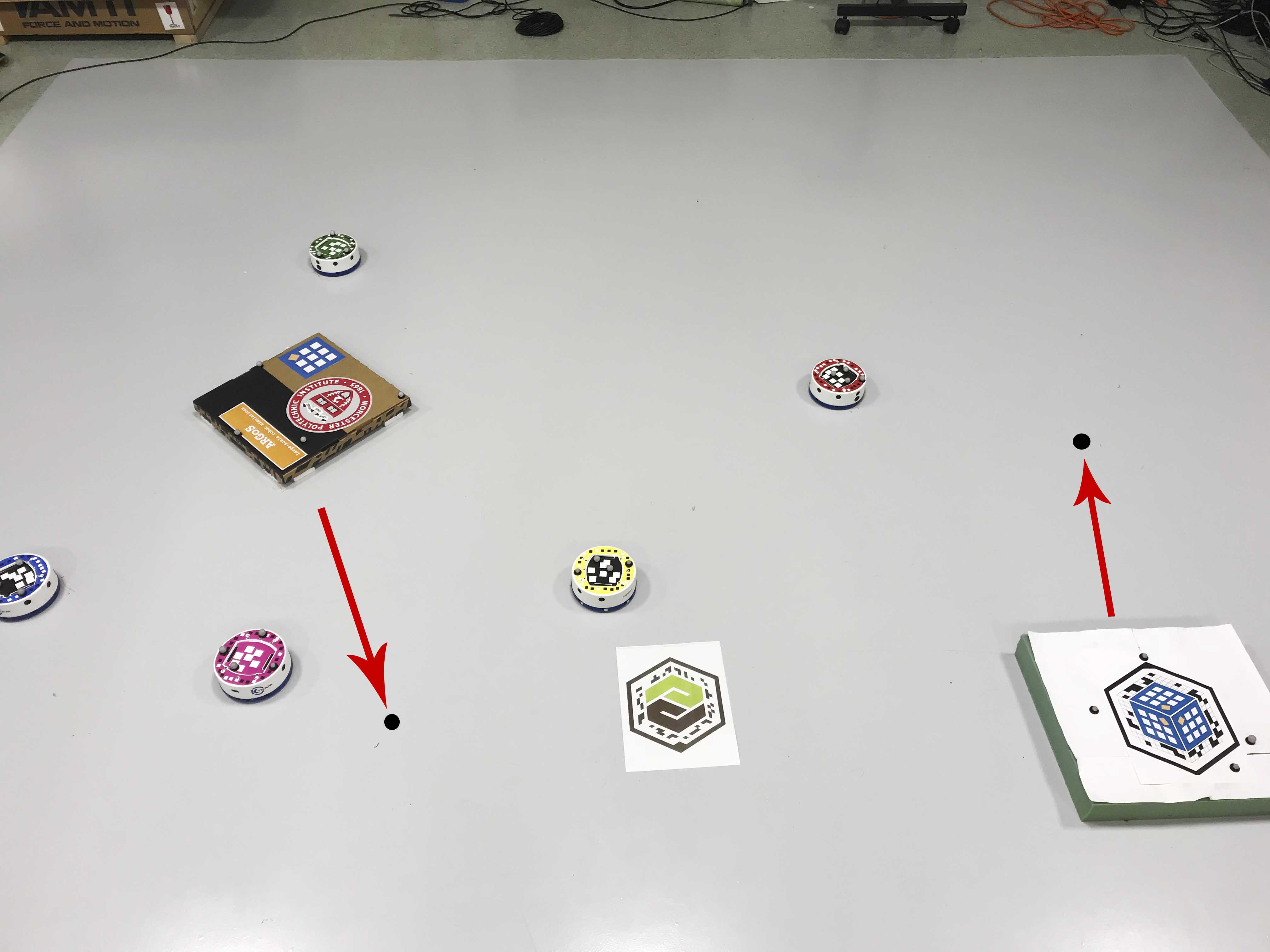}
            \caption{The setup of Experiment 2. The overlaid red arrows and black points indicate the directions and positions of the objects to be transported by the robots.}
            \label{fig:exp2}
        \end{figure}
        
        \textbf{Experiment 2: Setup.} We performed 10 consecutive trials of collective transport of two objects, Obj1 and Obj2, with five robots (see Fig.~\ref{fig:exp1}). Each object had to reach a predefined pose ($x_1 = \unit[0.8]{m}$, $y_1 = \unit[0]{m}$, $\theta_1 = \unit[128]{^\circ}$) and ($x_2 = \unit[-1]{m}$, $y_2 = \unit[0.5]{m}$, $\theta_2 = 46^{\circ}$) respectively. We performed all the trials with a human-in-the-loop to reassign the robots to Obj2 upon completion of Obj1's transport. 
        

        \textbf{Experiment 2: Results.} Table~\ref{tab:exp2} shows the statistics of position and orientation errors collected at the end of 10 trials. As Obj2 had only one assigned robot, the task stayed idle until Obj1 reached its goal pose and sufficient robots were reassigned by the user to Obj2. We kept track of the time at which the human interacts through the interface. The median time we observed was \unit[171.65]{s}, with a minimum of \unit[143.9]{s} and a maximum of \unit[244.5]{s}.
         These values are greater than the completion time of Obj1. This indicates that Obj2 was usually transported after the human interacted with the system. Hence, in most cases, Obj2 would not have been transported to the destination without human intervention.
	    
	    \begin{table}[t]
            \centering
            \caption{Statistics on data collected from Experiment 2, two object transport with human user's presence.}
            \begin{tabular}{c|c||c|c|c}
            \hline
            Object ID & Data type (unit) & Min & Max & Median \\
            \hline
                \multirow{4}{*}{Obj1} & Error on $x_1$ (m) &	            -0.0709	&	0.0525	&	0.0367	\\
                                      & Error on $y_1$ (m) &	         0.0130	&	0.1022	&	0.0493	\\
                                      & Error on $\theta_1$ ($^{\circ}$) &-0.9130	&	1.2540	&	0.8635\\
                                      & Completion Time (s) &	 134.20	&	232.10	&	159.35\\
            \hline
                \multirow{4}{*}{Obj2} & Error on $x_2$ (m)	&	-0.0506	&	0.00954	&	0.0012	\\
                                & Error on $y_2$ (m)	&	-0.1171	&	-0.0435	&	-0.0862	\\
                                & Error on $\theta$ ($^{\circ}$)	&	-0.8786	&	1.1758	&	0.6672	\\
                                &   Completion Time (s)	&	295.9	&	402.1	&	363.3	\\
                                    
            \hline
            \end{tabular}
            \label{tab:exp2}
        \end{table}  
        
    
    
    \subsection{Usability  Analysis}
        
        \begin{figure}[t]
          \centering
            \includegraphics[width=0.4\textwidth]{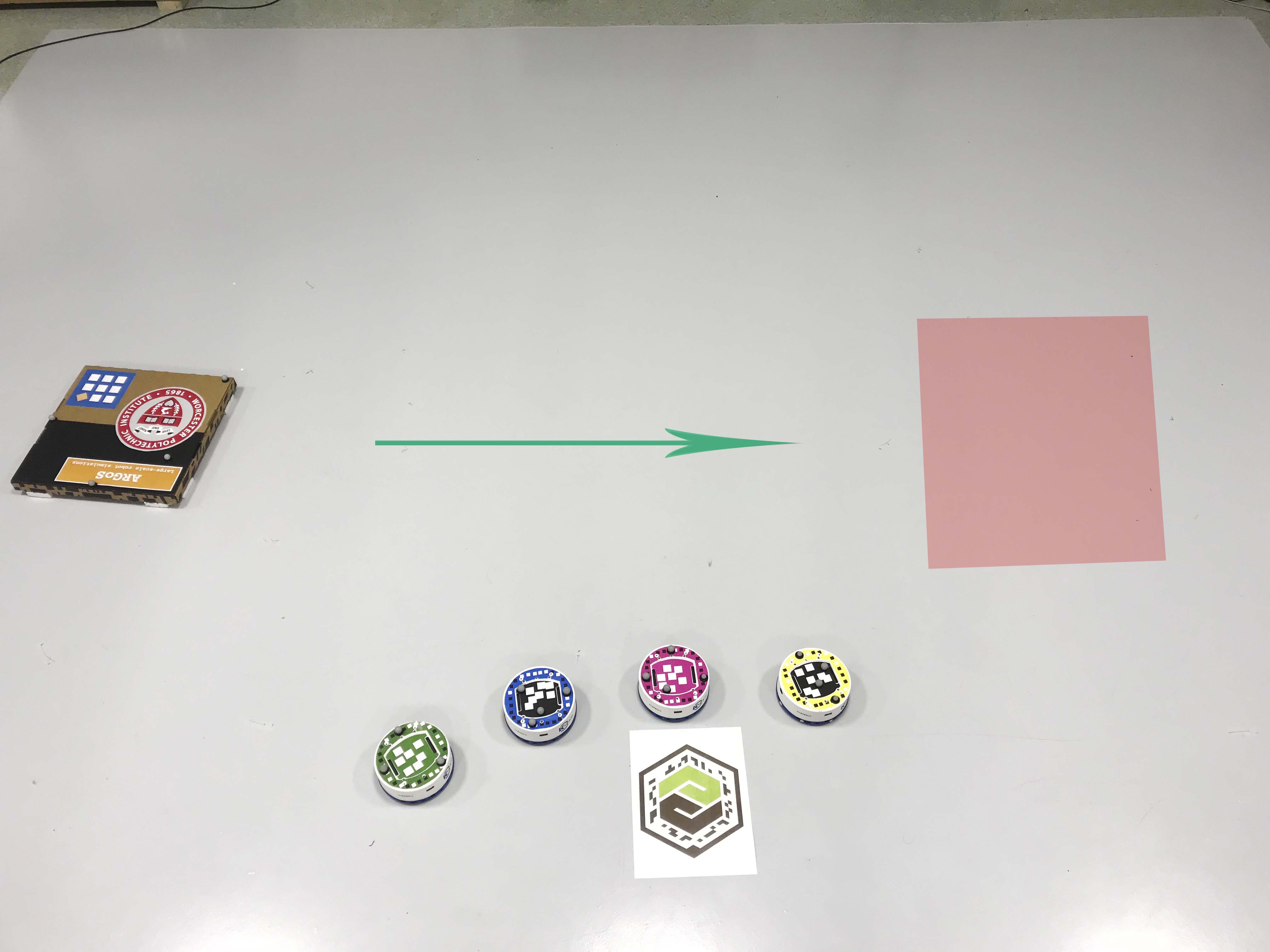}
          \caption{User study experimental setup. The overlaid green arrow indicates the direction of the object to be transported by the robots. The overlaid red region indicates the goal region in which the object needs to be translated.}
            \label{fig:ust}
        \end{figure}
        
        \begin{figure*}[t]
          \centering
            \includegraphics[width=.8\textwidth]{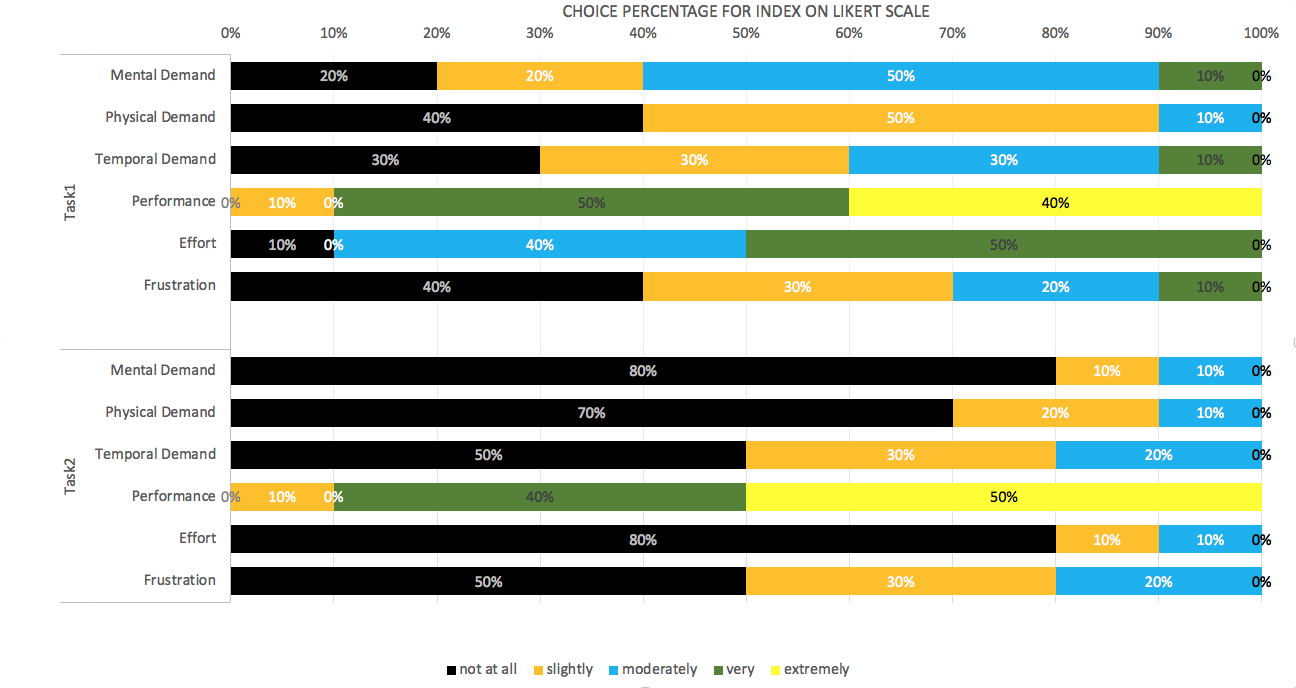}
          \caption{Percentage of raw scores using NASA TLX subscales on the Likert scale from the questionnaire submitted by users after the study.}
            \label{fig:rawscore}
        \end{figure*}
        
        \begin{figure}[t]
          \centering
            \includegraphics[width=0.4\textwidth]{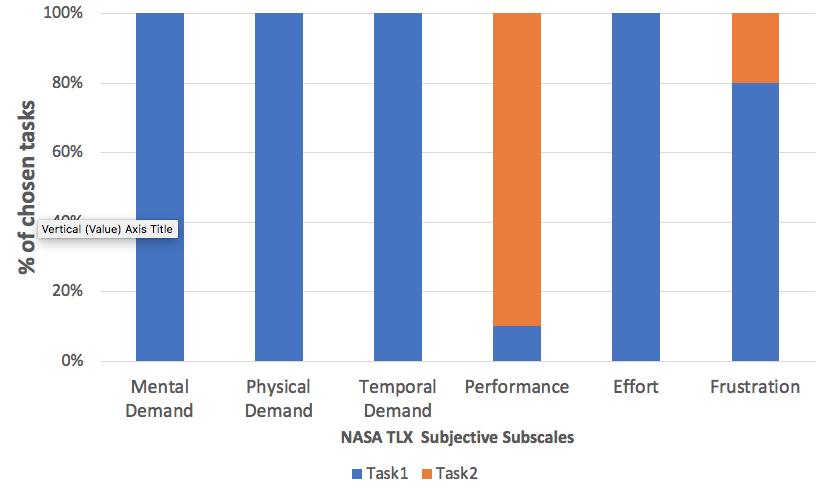}
          \caption{Comparative cognitive load.}
            \label{fig:compare}
        \end{figure}
    
        \textbf{Experimental Setup.} We conducted a user study of 10 participants from multiple disciplines. The users' ages ranged between 20 and 29. The experiment in which every user was involved was divided into two halves, with one task to complete for each half. After each half, the participants were asked to fill out a questionnaire based on NASA TLX~\cite{hart_development_1988} subscales using a Likert Scale~\cite{likert} for quantifying their response. All the participants had no prior experience of interacting with the system. 
        
        \textbf{Task Information.} Both tasks involved transporting an object to the goal region (Fig.~\ref{fig:ust}) using our interface. The participants had to focus on only translation of the object and not the orientation. The task ended when the object completely entered the goal region. In Task1, the participants could only control up to four robots manually (a purely robot-oriented interaction). In Task2, the participants could interact with the object (environment-oriented) and with the robots (robot-oriented).
        
        \textbf{Results.} To quantify the reactions of the participants during the experiments, we employed the NASA TLX scales on a Likert Scale and the results are shown in Fig.~\ref{fig:rawscore}. The percentage in the plot depicts the number of responses made for a particular scale index. Fig.~\ref{fig:compare} reports the results of the comparative study, where the participants were asked to indicate which task caused a more significant cognitive load. The comparative study clearly shows that Task2, in which both environment- and robot-oriented interactions were allowed, has a lower cognitive load with respect to Task1, which only allowed robot-oriented interactions. The results of Fig.~\ref{fig:compare} confirm the raw scores displayed in Fig.~\ref{fig:rawscore}. Further evidence supporting our claim that combining environment- and robot-oriented interactions is beneficial is provided by the number of interactions recorded during the experiments. The data is reported in Table~\ref{tab:interactions}, and it clearly shows that combining the two modalities entails a lower number of operations than purely robot-oriented approaches.
        
        \begin{table}[t]
            \centering
            \caption{Statistics on number of interactions made with the interface by the user in both the task during the user study.}
            \begin{tabular}{c||c|c|c}
            \hline
            \multicolumn{1}{c||}{} & \multicolumn{2}{c}{\hfill No. of Interactions \hfill} \\
            \cline{1-4}
            Task ID & Min & Max & Median  \\
            
            \hline
                 Task 1              & 8   & 118 & 52      \\
                 Task 2              & 1   & 8   & 1       \\
            \hline
            \end{tabular}
            \label{tab:interactions}
        \end{table}   
        

        \section{Conclusion and Future Work} \label{sec:conclusion}
        In this paper, we propose a new type of human-swarm interface that combines environment-oriented and robot-oriented modalities of interaction. We based our interface on an app for a hand-held device, because of both the low cost of this solution and the intuitiveness of touch-based graphical interfaces. Our interface and the associated infrastructure are designed to work with simulated as well as real robots, although in this paper we focused our analysis on real-robot experiments. We performed a user study to validate the effectiveness of our interface. Results confirmed that, for a task such as collective transport, the ability to mix environment-oriented commands and robot-oriented ones is beneficial, because it results in a lower number of required commands to achieve the goal.

        In a broader perspective, this work suggests that the specifics of the task a robot swarm must complete, plays an important role in the definition of human-swarm interfaces. As discussed in~\cite{kolling_human_2013}, environment-oriented interactions might not scale well for tasks that involve diffusion of many robots in cluttered environments. However, for tasks such as collective transport, in which the robots are tightly connected to the object to carry, the ability to focus on the object makes the interaction more effective.

        In future work, we will explore several directions. First, we intend to perform a scalability study with simulated robots in conditions analogous to~\cite{kolling_human_2013} (multiple objects, cluttered environment, 50+ robots), with the aim of better comparing our findings with~\cite{kolling_human_2013}. Second, we will investigate how to efficiently express \emph{sequential} and \emph{parallel} tasks, for instance by developing a visual language to encode the relationships among the tasks. Third, we will investigate the role of \emph{interference} among parallel tasks on the usability of human-swarm interfaces.
    
    %


\bibliographystyle{IEEEtran}
\bibliography{refs,zotero}

\end{document}